\documentclass[cameraready]{Interspeech}

\usepackage{cite}
\usepackage{amsmath,amssymb,amsfonts}
\usepackage{algorithmic}
\usepackage{graphicx}
\usepackage{textcomp}
\usepackage[table]{xcolor}
\usepackage{hyperref}
\usepackage{booktabs}
\usepackage{multirow}

\renewcommand{\paragraph}[1]{\noindent\textbf{#1}\quad}

\title{\textsc{AnySimLite}: A Lightweight Few-Shot Similarity Encoder for On-Device Speech-Adjacent Classification}

\author[affiliation={1}, orcid=0000-0003-1866-1408]{Sourav}{Ghosh}
\author[affiliation={1}, orcid=0009-0005-0196-7621]{Yash}{Bhatia}
\author[affiliation={2}, orcid=0009-0007-6207-8080]{Keshav}{Goyal}
\author[affiliation={3}]{Sahil Singh}{Bagri}
\author[affiliation={1}, orcid=0000-0003-4312-7004]{Mohamed Akram Ulla}{Shariff}
\author[affiliation={1}, orcid=0009-0008-3912-2247]{Saravana Balaji}{Shanmugam}

\address{
    $^{1,2,3}$ Samsung R\&D Institute Bangalore, India
}

\email{sourav.ghosh@samsung.com, yash.bhatia@samsung.com, keshavgoyal885.kg@gmail.com, sahil.singhbagri.cse22@itbhu.ac.in, m.shariff@samsung.com, saravana.bs@samsung.com}


\usepackage{comment}

\begin{document}

\maketitle

\begin{abstract}
    To minimize privacy concerns and inference latency on edge devices like smartphones, lightweight on-device models remain important for end-user applications. Many of these applications involve natural language classification, but deploying multiple specialized models creates a memory footprint challenge. We investigate: \textit{Can a single lightweight architecture solve multiple Speech-Adjacent (SA) classification tasks through reduction to a nuanced text similarity formulation?} We propose \textsc{AnySimLite}, a lightweight similarity encoder that combines word-level and character-level channels. Together with a dataset transformation strategy, we evaluate \textsc{AnySimLite} across multiple SA classification tasks and show that it consistently achieves state-of-the-art (SOTA) or SOTA-competitive performance in few-shot settings while maintaining a low memory footprint. Even in the worst case, the performance drop remains below 7\% while using $<\frac{1}{250}^{\mathrm{th}}$ of the model size of the SOTA \mbox{qLLaMA\_LoRA-7B} baseline.
\end{abstract}

\footnotetext[2]{\textsuperscript{,3} These authors contributed to the work during their former employment and internship respectively.}

\section{Introduction}

On-device models are essential in inference pipelines on edge devices for obvious benefits in terms of network latency, data privacy, and overall low carbon footprint at data centers. These models should have low latency and low resource requirements (storage, memory, power). Modern smartphones contain multiple models as part of SDK runtimes and OEM/third-party applications. Due to the sheer number of such models, these add up in storage resource requirement. Thus, there is a perennial need to optimize the resource consumption for all on-device solutions. Now, many on-device models are designed for very specific use cases, and thus their architectures and training methodologies pertain only to narrow scenarios. However, there are many other redundant models across different solutions and SDKs that solve an identical core task or target problems that are reducible to the same problem.
Many speech related tasks require processing of transcriptions from automatic speech recognition (ASR). One salient example is intent detection in voice assistants, where transcripts need to be classified into a set of multiple known intents.
In speech-adjacent natural language processing (NLP), this involves text similarity.

\begin{figure}[t]
    \centering
    \includegraphics[width=\linewidth]{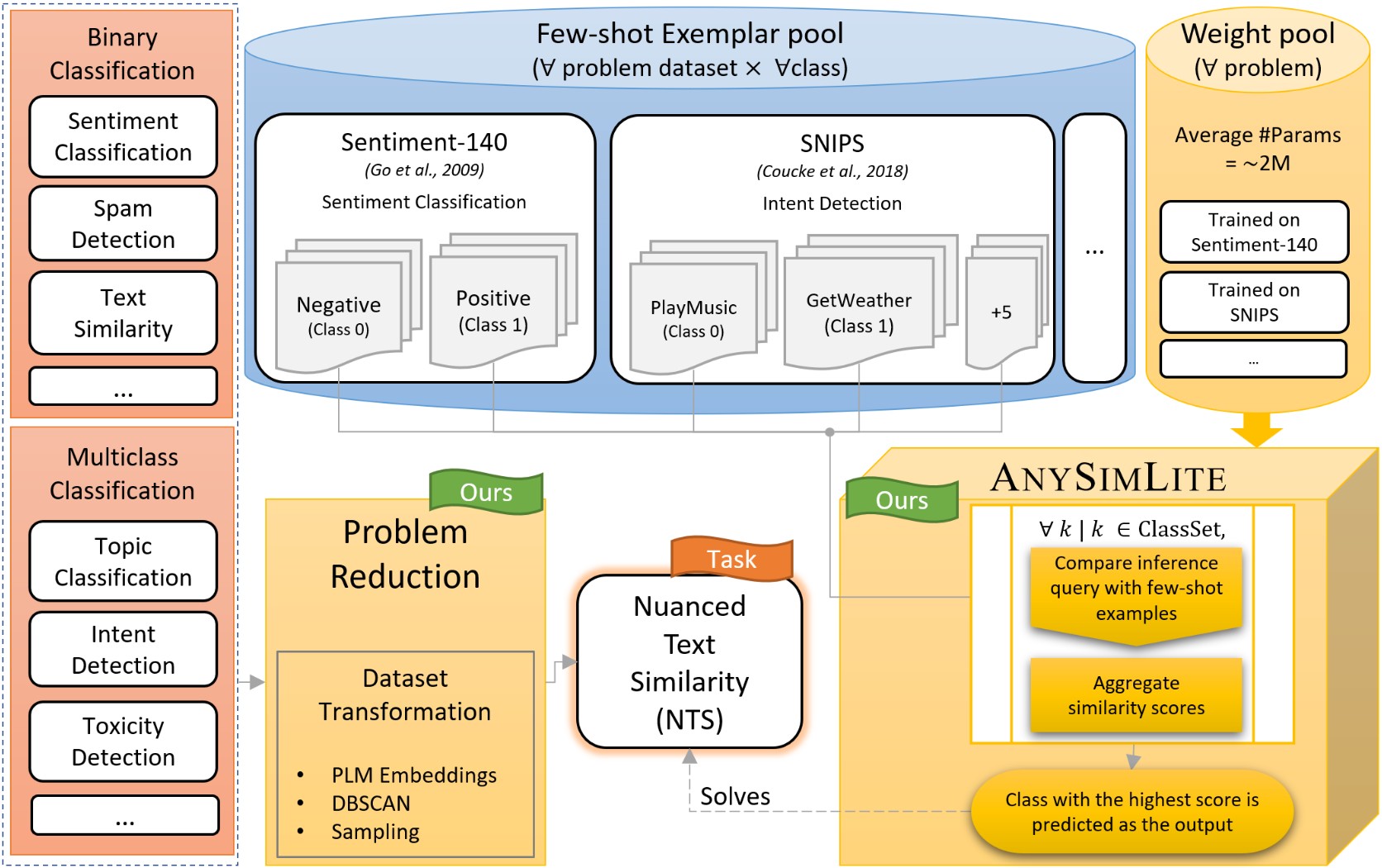}
    \caption{Solving NLP classification via reduction to NTS.}
    \label{fig:teaser}
\end{figure}

Text similarity is the task of identifying whether two text units are similar or not. The input covers short texts like tweets to long text documents. Text similarity may also cover quantifying the similarity between a short query and a much longer document. There are multiple approaches in literature towards text similarity, ranging from token matching and TF-IDF to semantic embeddings. In practice, based on the specific problem and the available dataset(s), one of these approaches is selected -- algorithmic, neural, or a hybrid of both. In many cases, the text similarity may also be highly specialized, where two texts are said to be similar solely on the basis of the alignment of certain common features and not in the traditional lexicographic or semantic sense. We refer to this as \textit{``nuanced text similarity''} (NTS).

To this end, we are motivated to design a lightweight architecture for on-device deployment to singularly address a multitude of tasks that are reducible to an NTS task (Fig.~\ref{fig:teaser}). At its core, we propose a set of lightweight architectures and tune them with a toy problem called \textit{``Event Title Similarity''}. The designs of these architectures are motivated by the goals of on-device deployment and near-real-time latency for concerned applications. We first showcase the efficacy of such architecture in solving the toy problem, and conduct detailed ablation study to identify the components with the most significant impact. Thereafter, we identify the best variant of each proposed architecture and evaluate their usefulness in addressing different problem statements on popular public-domain datasets using relevant metrics.

Our contributions can be summarized as:
\begin{itemize}
    \item We propose a lightweight architecture to solve an exemplar nuanced text similarity (NTS) problem titled \textit{``Event Title Similarity''}, optimized for on-device deployment and low inference latency.
    \item We demonstrate how multiple NLP tasks can be reduced to NTS and solved using our lightweight architecture in a few-shot setting with SOTA-competitive performance. Our proposed model \textsc{AnySimLite} consistently achieves state-of-the-art (SOTA) or SOTA-competitive scores with a significantly lower memory footprint.
    \item We introduce a novel dataset transformation technique to derive labeled pairs of documents from classification datasets with focus on sampling \textit{``hard''} pairs.
\end{itemize}

To the best of our knowledge, we are the first to explore a unified lightweight model for solving a variety of speech-adjacent NLP classification tasks via problem reduction.

\section{Preliminaries}

\subsection{Problem Formulation}\label{sec:problem_formulation}
In text similarity problem, the aim is to model the similarity between two text documents to a quantifiable score. Mathematically, the goal is to formulate a model $f$, such that two text documents, $d_1$ and $d_2$, are said to be more similar, compared to $d_3$ and $d_4$, if and only if $f\left(d_1, d_2\right) > f\left(d_3, d_4\right)$. This implies that the commutative property holds, i.e., $f\left(d_i, d_j\right) = f\left(d_j, d_i\right)$.

In this work, we demonstrate a methodology to reduce select NLP problems to a nuanced text similarity (NTS) task. Thus, for a task $\mathcal{G} \in \mathbf{R} \subset \mathbf{L}$, where $\mathbf{L}$ is the set of all NLP tasks and $\mathbf{R}$ is a subset of tasks that are reducible to NTS, we convert $\mathcal{G}$ to $\mathcal{G'}$, such that $\mathcal{G'} \equiv f$. Thereafter, by solving $\mathcal{G}'$, we find a solution to $\mathcal{G}$.

\subsection{Description of Toy Problem}
In this section, we describe our toy problem statement and the reason for its selection. Text similarity has primarily been explored in literature with the aim to ascertain whether two text documents are semantically similar or not. Whereas na\"ive approaches to this may involve token matching, modern approaches involve encoding the document into an embedding that captures the semantic attributes of the documents. However, a more generalized scenario would be where the similarity of documents is not confined to either of these two extremes, i.e., where similarity does not imply semantic similarity only. For instance, it may be of interest to know whether two product reviews on an e-commerce websites are similar in terms of sentiment even if there is a significant difference among the reviewers' experiences behind the positive or negative sentiment. Similarly, two emails may be considered similar based on whether both of them are spam or not, regardless of their actual contents. This opens the possibility of solving a variety of problems like sentiment classification and email spam detection simply by computing the similarity of a query document to a set of pre-annotated exemplars from each corresponding class in a few-shot setting.

To develop a foundation for this nuanced text similarity, we need to consider a problem where a pair of text documents have a unique constraint for being considered as similar or not similar. In order to minimize biases in the architecture, such a constraint has to be non-obvious. Furthermore, since our motivating use case is on-device deployment, we strive to select a set of data points that are readily available on an end-user device. We thus indulge a toy problem for NTS called \textit{``Event Title Similarity''}, where two event titles are said to be similar if and only if the corresponding two text units describe the same underlying events and deal with the same set of people involved. For instance, \textit{``Birthday party for John''} is to be considered similar to neither \textit{``Meeting with John''} nor \textit{``Sarah's birthday party''} because, in each case, at least one of the underlying event category or the concerned named entities (NEs) differ. Thus, in this case,

\begin{equation}
    f\left(d_1, d_2\right) \equiv \eta\left(d_1, d_2\right) \wedge \nu\left(d_1, d_2\right)
    \label{eq:event_title_similarity_function}
\end{equation}

where $\eta$ and $\nu$ denote functions classifying whether titles $d_1$ and $d_2$ deal with the same event and same NEs respectively. For complexity analysis, we consider the problem when a database (DB) or knowledge graph (KG) is to be populated with event titles without duplication of similar titles. The scenarios are -- (a) Initialization or \texttt{Init}: when event titles are loaded into an empty DB/KG, and (b) Update or \texttt{Add}: when an already populated DB/KG is to be updated with one additional event title.

\begin{figure}[t]
    \centering
    \includegraphics[width=\linewidth]{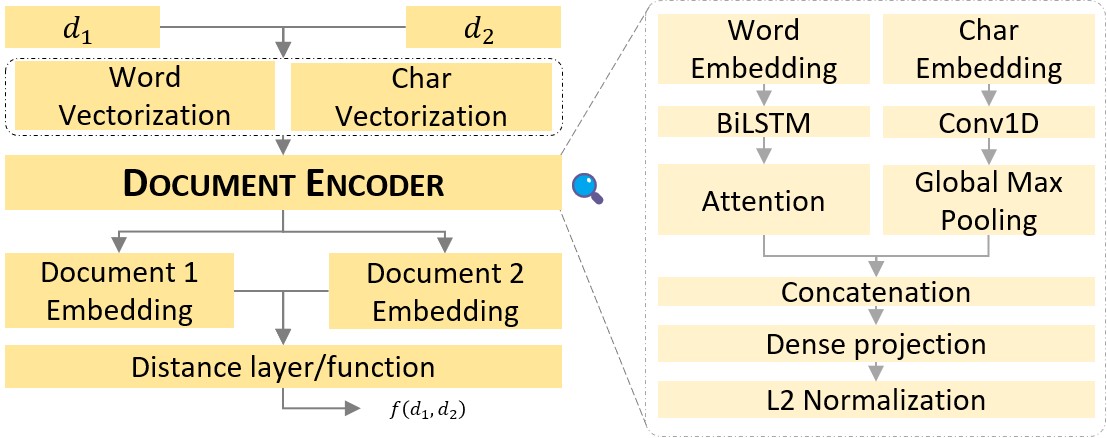}
    \caption{\textbf{Architecture of \textsc{AnySimLite}} consisting of a lightweight encoder with word and character channels.}
    \label{fig:architecture}
\end{figure}

\begin{table*}[t]
    \centering
    \resizebox{\linewidth}{!}{
        \begin{tabular}{c|p{0.15\linewidth}|c|c|c|c|c|c|c}
            \toprule
            \textbf{Treatment} &   \textbf{Dataset}& \textbf{Alias} & \textbf{Configuration} & \textbf{\#Params (M)} & \textbf{Precision} & \textbf{Recall} & \textbf{Accuracy} & ${F}_1$ \\\midrule
            Binary&   Labelled&A1&Word embeddings w/ BiLSTM& {0.42}& {55.21}& {77.28}& {62.34}& {64.40}\\
            Classification&   Pairs&A2&A1 + Char embeddings w/ Conv1D& {0.42}& {77.35}& {76.30}& {76.32}& {76.82}\\\midrule
            \multirow{6}{*}{Enc}&   &B1&Sentence Transformer & {22}& {87.59}& {86.11}& {86.14}& {86.84}\\
            &   Labelled&B2&A2& {0.42}& {80.46}& {80.91}& {80.42}& {80.68}\\
 &  Pairs& \cellcolor{gray!25} B3& \cellcolor{gray!25} B2 + Attention@word (\textsc{AnySimLite} Base)&  \cellcolor{gray!25} {0.42}&  \cellcolor{gray!25} {88.07}&  \cellcolor{gray!25} {90.41}&  \cellcolor{gray!25} {88.73}& \cellcolor{gray!25} {89.22}\\
 &  &B4& B3 + Attention@char& {0.49}& {84.24}& {85.77}& {85.43}&{84.99}\\
 &  &B5& B2 + Cross-channel attention& {0.54}& {81.27}& {85.39}& {84.88}&{83.27}\\\cline{2-9}
 &   Triplets&B6&B3 + Siamese Network& {0.42}& {80.44}& {71.19}& {79.36}&{75.53}\\\midrule
 Clustering&   Cluster labels&B7&B3 + Origin Clustering& {0.63}& {81.64}& {84.11}& {83.32}&{82.85}\\\midrule
 \rowcolor{gray!25} KD& As in B3 & B8 & B3 + MiniLM L12-v2 (\textsc{AnySimLite} Deployment variant) & {0.72}& {92.79}& {88.79}& {90.83}&{90.75}\\
 \bottomrule
        \end{tabular}
    }
    \caption{\textbf{Ablation study:} We select B3 as the optimal base candidate for \textsc{AnySimLite} architecture, using Word channel with Attention layer and Char channel with Global Max Pooling and simple Concatenation, and B8 as its distilled deployment variant.}
    \label{tab:ablation}
\end{table*}

\section{Architecture for NTS}
We describe configs. explored to solve the toy problem, and thereby, to act as foundation for lightweight on-device NTS.

\subsection{As binary classification}
By concatenating the two titles together, a single string is produced that can be tokenized and fed to the model architecture. Here, a training dataset for supervised learning is to be organized into pairs of titles with similarity labels in binary.

\subsubsection{Word embeddings}
Using only word embeddings enables a very low memory footprint, but due to vocab limitations, such a model is unable to identify differences in out-of-vocab (OOV) NEs which is a key requirement. While our toy problem has less dependency on word token ordering, to generalize across downstream NLP tasks, we feed the word embeddings to a BiLSTM layer. \texttt{Init}, \texttt{Add} latency is of $\mathcal{O}\left(n^2\right)$, $\mathcal{O}\left(n\right)$.

\subsubsection{Word and character embeddings}
To solve the above problem with OOV NEs, we introduce an additional channel of character embeddings. \texttt{Init}, \texttt{Add} latency is of $\mathcal{O}\left(n^2\right)$, $\mathcal{O}\left(n\right)$.

\subsection{As title encoder}
One key issue with concatenating titles as in binary classification is that they violate commutative property of text similarity (refer to Section \ref{sec:problem_formulation}). To resolve this, we transition to an encoder network approach that creates a unique embedding for a title. The similarity of two titles can then be the output of cosine similarity of the two corresponding embeddings.

\subsubsection{Sentence Transformers}
Using sentence transformers in both pretrained and fine-tuned settings reduces training data requirement. It further leads to a latency of $\mathcal{O}\left(n\right)$ and $\mathcal{O}\left(1\right)$ for \texttt{Init} and \texttt{Add} respectively.

\subsubsection{Word and character embeddings -- \textsc{AnySimLite}}
To reduce the memory footprint of the encoder, we replace the sentence transformer architecture with a simple network comprising of the word and character embedding channels. This approach preserves the benefits of an encoder architecture, including low latency ($\mathcal{O}\left(n\right)$ and $\mathcal{O}\left(1\right)$ for \texttt{Init} and \texttt{Add} respectively) and also that of identifying OOV NE differences due to the presence of character embeddings along side word embeddings.

Based on empirical results and ablation study, we select this as Event Title Similarity architecture and refer to this model as \textsc{AnySimLite}. Fig. \ref{fig:architecture} details the architecture of this model.

\subsubsection{Notable alternatives}

\paragraph{As Siamese network.}
We experiment with using a Siamese network with triplet-based training process instead of pairs of titles. However, the challenge of sampling hard examples of Anchor-Negative (dissimilar but not-too-dissimilar) requires hand-crafting and domain knowledge of the problem statement. Since our aim is to ultimately use the architecture as a foundation for multiple tasks, this approach is infeasible due to the aforesaid challenge.
\\

\noindent\paragraph{As origin clustering.}
By organizing training data with clusters of similar titles, we experiment with a multiclass classification approach, where each class effectively denotes a combination of one event and a set of NEs. While this approach works well with a validation set following the same data distribution and demographics of NE (many common events and NEs), it does poorly when event titles in test set do not follow the same distribution. Thus, we discard this.

\begin{table*}[t]
    \centering
    \resizebox{0.81\linewidth}{!}{
        \begin{tabular}{c|c|c|c|c|r}
            \toprule
            \textbf{Task} $\mathbf{\in} \mathbf{R}$ & \textbf{Dataset} & \textbf{Metric} & \textbf{Method} & \textbf{Score ($\Delta$ w.r.t. best)} & \textbf{\#Params (M)} \\\midrule
    
            \multirow{8}{*}{Text Similarity}          &                                                    & \multirow{3}{*}{Acc}         & BERT & 75.39 & 108 \\
                                                      & \multirow{1}{*}{TitleSimCurated}                   &                              & MiniLM L12-v2 & 84.16 & $\sim$25 \\
                                                      & \multirow{1}{*}{\footnotesize Classes = 2}                &                              & MiniLM L6-v2 & 86.14 & $\sim$20 \\
                                                      &                                                    &                              & \cellcolor{gray!25} \textsc{AnySimLite} & \cellcolor{gray!25} \textbf{88.73} (3.01\% $\uparrow$) & \cellcolor{gray!25} \textbf{0.1}$^{*}$ \\\cline{2-6}
                                                      &                                                                     & \multirow{3}{*}{${F}_1$/Acc} & Sharma et al., 2019 \cite{sharma2019naturallanguageunderstandingquora} & 66.30 / 74.60 & - \\
                                                      &                                                                     &                                & S-CNN \cite{han2022building} & 82.02 / 83.32 & - \\
                                                      & \multirow{1}{*}{Quora Question Pairs \cite{question_pairs_dataset}} &                                & qLLaMA\_LoRA-7B \cite{han2025enhancing} & \textbf{84.90} / \textbf{88.67} & 7000 \\
                                                      & \multirow{1}{*}{\footnotesize Classes = 2}                                 &                                & LLaMA-33B\_5shot \cite{han2025enhancing} & 49.39 / 63.36 & 33000 \\
                                                      &                                                                     &                                & \cellcolor{gray!25}\textsc{AnySimLite} & \cellcolor{gray!25} 82.96 / 82.77 (2.29\% $\downarrow$ / 6.65\% $\downarrow$) & \cellcolor{gray!25} \textbf{2.6} \\\midrule
            
            \multirow{8}{*}{Sentiment Classification} &                                                           & \multirow{3}{*}{${F}_1$/Acc} & RoBERTa-BiLSTM \cite{rahman2025roberta} & \textbf{82.25} / 82.25 & \textgreater 125 \\
                                                      & \multirow{1}{*}{Sentiment-140\cite{go2009twitter}}        &                         & BERT \cite{elmitwalli2024sentiment} & 81.14 / \textbf{82.54} & 110 \\
                                                      & \multirow{1}{*}{\footnotesize Classes = 2}                       &                         & GPT-3 \cite{elmitwalli2024sentiment} & 79.13 / 79.11 & 175000 \\
                                                      &                                                           &                         & \cellcolor{gray!25} \textsc{AnySimLite} & \cellcolor{gray!25} 80.17 / 80.22 (2.53\% $\downarrow$ / 2.81\% $\downarrow$) & \cellcolor{gray!25} \textbf{1.3} \\\cline{2-6}
                                                      &                                                                       & \multirow{3}{*}{${F}_1$/Acc} & RoBERTa-BiLSTM \cite{rahman2025roberta} & 92.35 / 92.36 & \textgreater 125 \\
                                                      & \multirow{1}{*}{IMDB Movie Reviews \cite{maas-EtAl:2011:ACL-HLT2011}} &                         & BERT \cite{elmitwalli2024sentiment} & \textbf{93.80} / \textbf{93.80} & 110 \\
                                                      & \multirow{1}{*}{\footnotesize Classes = 2}                                   &                         & GPT-3 \cite{elmitwalli2024sentiment} & 91.19 / 90.76 & 175000 \\
                                                      &                                                                       &                         & \cellcolor{gray!25} \textsc{AnySimLite} & \cellcolor{gray!25} 88.68 / 88.68 (5.46\% $\downarrow$ / 5.46\% $\downarrow$) & \cellcolor{gray!25} \textbf{1.1} \\\midrule
    
            \multirow{6}{*}{Intent Detection}         &                                                                           & \multirow{3}{*}{Acc} & ESIE-BERT \cite{guo2023esiebertenrichingsubwordsinformation} & \textbf{99.10} & \textgreater 110 \\
                                                      & \multirow{1}{*}{SNIPS \cite{coucke2018snipsvoiceplatformembedded}}        &                      & LIDSNet \cite{9680131} & 98.00 & 0.59 \\
                                                      & \multirow{1}{*}{\footnotesize Classes = 7}                                       &                      & \cellcolor{gray!25} \textsc{AnySimLite}    & \cellcolor{gray!25} 98.21 (0.90\% $\downarrow$) & \cellcolor{gray!25} \textbf{0.35} \\\cline{2-6}
                                                      &                                                                    & \multirow{3}{*}{Acc} & ESIE-BERT \cite{guo2023esiebertenrichingsubwordsinformation} & \textbf{98.10} & \textgreater 110 \\
                                                      & \multirow{1}{*}{ATIS \cite{price-1990-evaluation}}                 &                      & LIDSNet \cite{9680131} & 95.97 & \textbf{0.065} \\
                                                      & \multirow{1}{*}{\footnotesize Classes = 8 (coarse), 21 \cite{9680131}}      &                      & \cellcolor{gray!25} \textsc{AnySimLite} & \cellcolor{gray!25} 97.62 (0.49\% $\downarrow$) & \cellcolor{gray!25} 0.12 \\\midrule
    
            \multirow{3}{*}{Spam Detection}           &                                                                           & \multirow{3}{*}{${F}_1$/Acc} & Shen et al., 2025 \cite{SHEN202579} & 97.08 / \textbf{99.28} & \textgreater 110 \\
                                                      & \multirow{1}{*}{SMS Spam Collection\cite{sms_spam_collection_228}}        &                         & Liu et al., 2021 \cite{9433507} & 96.13 / 98.92 & - \\
                                                      & \multirow{1}{*}{\footnotesize Classes = 2}                                       &                         & \cellcolor{gray!25} \textsc{AnySimLite} & \cellcolor{gray!25} \textbf{97.50 / 99.28} (0.43\% $\uparrow$ / $=$) & \cellcolor{gray!25} \textbf{0.47} \\
                                                      \midrule
    
            \multirow{3}{*}{Topic Classification}     &                                                          & \multirow{3}{*}{Acc} & BERT-base + PGKD \cite{dipalo2024performanceguidedllmknowledgedistillation} & 89.50 & - \\
                                                      & \multirow{1}{*}{AG News \cite{NIPS2015_250cf8b5}}        &                      & Yang et al., 2019 \cite{yang2019xlnetgeneralizedautoregressivepretraining} & \textbf{95.50} & - \\
                                                      & \multirow{1}{*}{\footnotesize Classes = 4}                      &                      & \cellcolor{gray!25} \textsc{AnySimLite} & \cellcolor{gray!25} 91.12 (4.59\% $\downarrow$) & \cellcolor{gray!25} 1.3 \\
                                                      \midrule
    
            \multirow{3}{*}{Toxicity Detection}       &                                                                                     & \multirow{3}{*}{ROC}  & Kohli et al., 2018 \cite{kohli1184paying} & 72.40 & - \\
                                                      & \multirow{1}{*}{Toxic Comment \cite{jigsaw-toxic-comment-classification-challenge}} &                       & Chakrabarty, 2019 \cite{chakrabarty2019machine} & 73.00 & - \\
                                                      & \multirow{1}{*}{\footnotesize Classes = 6}                                          &                       & Toxic Crusaders \cite{jigsaw-toxic-comment-classification-challenge} & \textbf{98.86} & - \\
                                                      &                                                                                     &                       & \cellcolor{gray!25} \textsc{AnySimLite} & \cellcolor{gray!25} 94.20 (4.71\% $\downarrow$) & \cellcolor{gray!25} 3.6 \\
                                                      \bottomrule
        \end{tabular}
    }
    \caption{Empirical results demonstrate that a varied array of NLP tasks can be effectively reduced to an NTS task and solved by \textsc{AnySimLite} while achieving a SOTA-competitive performance metric with a low model size. (${}^*$Smaller embedding table due to reduced vocabulary.)}
    \label{tab:sota_comparison}
\end{table*}

\section{Datasets and Transformation}

\begin{figure}[t]
    \centering
    \includegraphics[width=\linewidth]{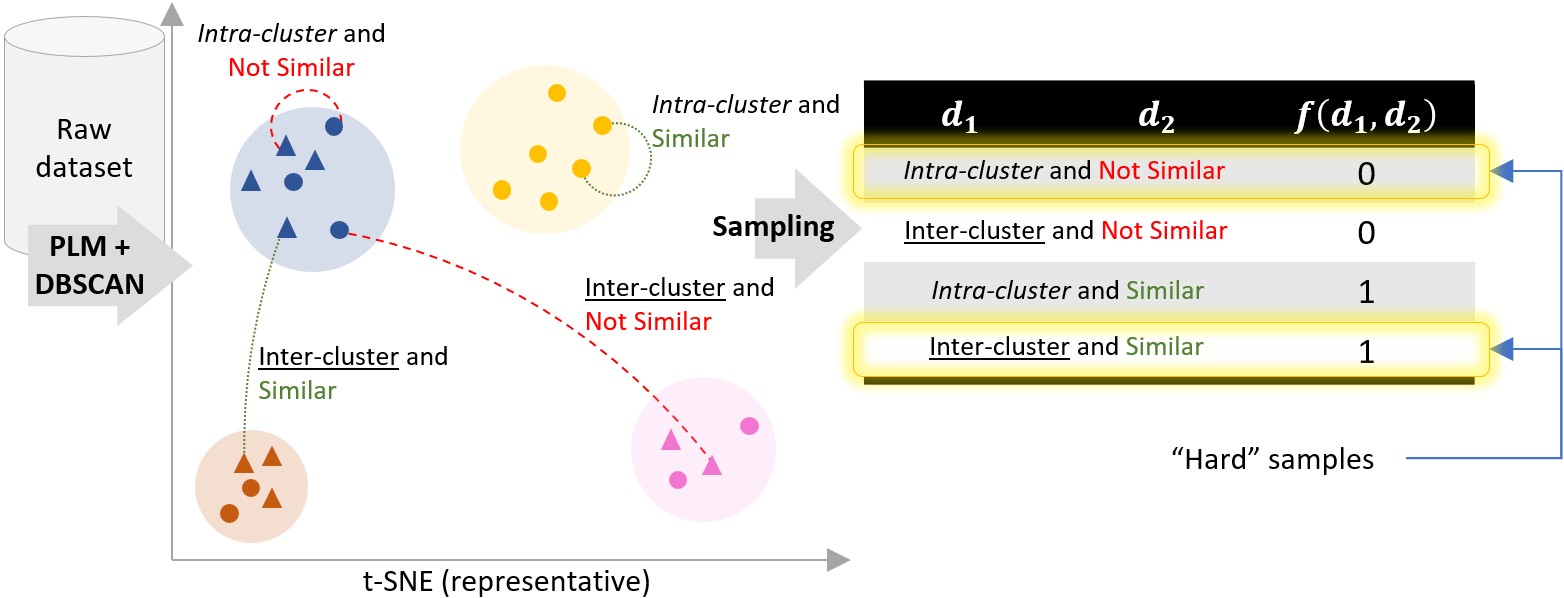}
    \caption{Dataset transformation}
    \label{fig:dataset-transformation}
\end{figure}

Recall from Section \ref{sec:problem_formulation} that $\mathbf{R}$ denotes a subset of tasks that are reducible to NTS. 
Once we have a foundation architecture for NTS in the form of \textsc{AnySimLite}, the next step is to devise a strategy to convert one of these tasks $\mathcal{G}$ to its NTS-reduced form, $\mathcal{G'}$. In this section, we introduce a few of such tasks in $\mathbf{R}$ along with their public-domain datasets, and briefly explain how each dataset $\mathcal{D}$ is transformed into $\mathcal{D'}$, containing pairs of documents for compatibility with our architecture training pipeline.

\subsection{Sampling of \textit{``hard''} pairs}
Curating pairs of samples from $\mathcal{D}$ at random for the transformed dataset would be a na\"ive approach and would lead to a high proportion of dissimilar samples which are \textit{``too dissimilar''}, preventing \textsc{AnySimLite} from learning the importance of the problem-specific nuance (like sentiment for sentiment analysis) in determining their dissimilarity. To tackle this challenge and identify hard samples, we first use an association metric to determine whether two samples are \textit{``too similar''}, \textit{``too dissimilar''}, etc. Using token-matching for this association metric would necessitate a $N \times N$ comparison of samples making it computationally expensive. On the other hand, using a TF-IDF approach in a rich vocabulary dataset would require a tremendous amount of memory to store the term-document matrix or sparse matrix optimization techniques.

Instead, we use a pretrained language model (PLM) to form embeddings of the documents in the dataset and then cluster them using DBSCAN. We then select intra-cluster and inter-cluster dissimilar samples at a constant ratio as shown in Fig. \ref{fig:dataset-transformation}. This is to ensure that a significant portion of the dissimilar samples are not \textit{``too dissimilar''} (their belonging to the same cluster implies shared factors notwithstanding the nuance specific to the problem statement). For our experiments, this ratio is 8:2.

\subsection{Selected problem statements $\in \mathbf{R}$}
We select diverse NLP classification tasks and their corresponding datasets, covering both binary and multiclass classifications for evaluating \textsc{AnySimLite} (Table \ref{tab:sota_comparison}). We also curate a dataset for our toy problem.

\subsubsection{TitleSimCurated dataset}
We curate a dataset specifically for ``Event Title Similarity'', containing two input strings and one binary label, compliant with Eq. \ref{eq:event_title_similarity_function}. This dataset is partially generated using hand-crafted templates and pretrained LLM with prompt engineering. It consists of 14 event categories -- anniversary, birthday, wedding, accommodation, travel, appointment, business, restaurant, social, activities, party, sports, graduation, and transportation.

\section{Experimental Results}
We conduct all training and experiments on an NVIDIA RTX A6000 GPU with 48 GB memory.

\subsection{Ablation Study}
To support the dual goal of \textsc{AnySimLite} architecture to be lightweight along with being versatile, we evaluate performance metric impact due to each component. For this purpose, we use TitleSimCurated dataset. From Table \ref{tab:ablation}, we note an optimal performance with B3 before Knowledge Distillation (KD). We attribute its superiority over B4 and B5 to the importance of attention layer in identifying key word tokens to focus on, which are not available to the cross-channel attention in B5. While intuitively B6 and B7 may be expected to showcase performance improvements, we attribute the lack of empirical backing to shortcomings in the dataset.

\subsection{Comparison with SOTA}

To evaluate our hypothesis that \textsc{AnySimLite} can solve multiple NLP classification tasks by reducing them to an NTS problem, we select 5 NLP tasks apart from Text Similarity and benchmark the performance of our model against their state-of-the-art (SOTA) approaches in a few-shot setting using 20 exemplar samples per class (see Fig.~\ref{fig:teaser}). Table \ref{tab:sota_comparison} shows that \textsc{AnySimLite} consistently achieves SOTA-competitive scores on all of these tasks. Notably, in spite of this high performance, our proposed model consistently has the smallest model size in all but one dataset. Furthermore, it outperforms the SOTA $F_1$ and accuracy metrics on SMS Spam Collection and accuracy metric on TitleSimCurated.

\subsection{On-device metrics}
We deploy \textsc{AnySimLite} on a Samsung Galaxy S25 Ultra device with a 8-bit quantized model occupying approximately 700 KB on disk ($\sim$700K parameters) and an inference latency of \textless~30~ms. For the few-shot examples of each dataset, we store precomputed 16-dimensional embeddings, which in a \texttt{float8} setting, have a negligible footprint of only 320\,B per class (20 exemplars $\times$ 16\,B). Thus, to support $J$ tasks, memory footprint is expected to be $J \times 700$KB.

\section{Conclusion}
We explore the hypothesis that a lightweight architecture based on word+char channels can solve NLP classifications via task reduction. Our \textsc{AnySimLite} achieves SOTA or SOTA-competitive performance, with an average accuracy degradation of only $2.24\%\pm3.23\%$ (sample standard deviation) relative to the best reported result, on diverse problem statements and public datasets with the least memory footprint. Our dataset transformation strategy samples ``hard'' examples for effective dataset curation. This work also demonstrates that lightweight models can remain competitive with multi-million-parameter models on certain classical NLP tasks. The future progression of this work can include exploring the feasibility of this proposal beyond classification tasks.

\section{Generative AI Use Disclosure}
Apart from explicit description in the paper, usage of generative AI tools is limited to permitted re-formatting of tables.

\bibliographystyle{IEEEtran}
\bibliography{AnySimLite}

\end{document}